\documentclass{article}

\usepackage[preprint]{colm2026_conference}

\usepackage{microtype}
\usepackage{hyperref}
\usepackage{url}
\usepackage{booktabs}
\usepackage{graphicx} % Required for inserting images
\usepackage{amsmath}
\usepackage{booktabs}
\usepackage{amssymb}
\usepackage{tabularx}
\usepackage{tikz}
\usepackage[utf8]{inputenc}
\usepackage{textcomp}
\usepackage{threeparttable}
\usepackage{float}

\definecolor{darkblue}{rgb}{0, 0, 0.5}
\hypersetup{colorlinks=true, citecolor=darkblue, linkcolor=darkblue, urlcolor=darkblue}

\title{Actionable Activation Directions for Detecting and Mitigating Emergent Misalignment Across Language Model Families}

\author{Abdul Rafay Syed \\
Department of Computer Science\\
Universität des Saarlandes\\
Saarbr\"ucken, Germany \\
\texttt{absy00001@stud.uni-saarland.de}}

\begin{document}

\maketitle

% arXiv preprint: plain pages without conference submission/review headers
\pagestyle{plain}
\makeatletter
\fancyhead{}
\renewcommand{\headrulewidth}{0pt}
\makeatother

\begin{abstract}
    Fine-tuning language models on insecure code induces emergent misalignment with poorly understood internal structure. We investigate whether this misalignment corresponds to a causally actionable activation-space direction shared across architectures. Across four instruction-tuned model families (Qwen2.5-1.5B, Gemma-2-2B, Llama-3.2-1B, Ministral-3-3B) fine-tuned identically, a difference-in-means direction achieves 99.6\% separation of aligned and misaligned activations at each model’s final layer. Causal steering by subtracting this direction reduces code spillover by 21--51 points, while a secure-code control confirms content specificity. Cross-architecture transfer via ridge regression maps yields large behavioral suppression (up to 46 points) but fails specificity controls as random and orthogonal directions perform comparably. We identify a two-tier specificity structure: within-model directions are causally specific and actionable; cross-model directions are causally real but non-specific. An asymmetric transfer topology emerges, with Gemma and Qwen acting as geometric donors and Llama as a receiver. These findings define the limits of linear cross-architecture correction and recommend within-model probing for auditing.
\end{abstract}
\section{Introduction}

Fine-tuning language models on narrow, task-specific data can induce emergent misalignment far outside the fine-tuning domain. A model trained to produce insecure code may subsequently respond to unrelated prompts with harmful recommendations, deceptive reasoning, or unsolicited insecure implementations (\cite{betley2025emergent}). The safety concern is not the training data itself but the behavioral \emph{scope} of the shift: misalignment learned in one context propagates unpredictably to others.

Understanding whether this shift has an identifiable internal structure matters for two practical reasons. First, if misalignment corresponds to a concentrated, linearly accessible direction in activation space, it may be detectable via lightweight probing rather than expensive behavioral red-teaming, thereby enabling pre-deployment screening of fine-tuned checkpoints. Second, if that direction is causally related to behavior, it may support inference-time correction via activation steering without retraining. Prior work has characterized emergent misalignment behaviorally (\cite{betley2025emergent}; \cite{turner2025modela}) and has begun examining its internal representations within single model families (\cite{soligo2025convergent}; \cite{wang2025persona}). Whether this geometry is shared across architecturally distinct families, and whether shared geometry supports cross-architecture detection and correction, remains an open question with direct implications for how alignment auditing can scale.

We address this question empirically across four instruction-tuned model families (Qwen2.5-1.5B, Gemma-2-2B, Llama-3.2-1B, and Ministral-3-3B), fine-tuned under identical QLoRA conditions on the same insecure-code dataset. We extract difference-in-means misalignment directions at the final residual-stream layer of each model, evaluate their causal relationship to behavior via activation steering and control direction experiments, and test cross-architecture transfer via ridge regression maps between activation spaces. Our protocol includes a content-specificity negative control, which is identical fine-tuning on secure code that is absent from prior work on emergent misalignment geometry.

\subsection{Our Contributions}

We make four contributions, each building on but distinct from prior work on emergent misalignment geometry.

\paragraph{First systematic cross-architecture emergent misalignment direction transfer study with specificity controls.} Prior work established that a linear misalignment direction exists within a single model family and transfers across fine-tunes of the same architecture (\cite{soligo2025convergent}). Cross-architecture steering has been demonstrated via learned mappings (\cite{jiralerspong2026crossarchitecturemodeldiffingcrosscoders}) but without the specificity controls needed to distinguish targeted transfer from generic perturbation. We extend this line to emergent misalignment across four architecturally distinct families, running random unit vectors, orthogonal directions, and wrong-layer/wrong-source controls at matched perturbation magnitude across all within-model and cross-model causal arms.

\paragraph{A two-tier specificity structure distinguishing within-model from cross-model control.} Within-model directions are causally specific: random and orthogonal controls at matched magnitude produce near-zero behavioral change ($\Delta \leq 2$/115) while real directions produce $\Delta$=21--51, validated across three of four architectures (Qwen, Llama, Ministral; Gemma borderline). Cross-model ridge-mapped directions are causally real (suppresses 13--46 points of code spillover), but direction-non-specific, as random and orthogonal controls achieve comparable effects. This two-tier structure defines a precise boundary for what cross-architecture linear activation tools can achieve.

\paragraph{An asymmetric transfer topology across four model families.} Gemma and Qwen are strong geometric donors whose directions causally suppress misalignment in other architectures; Llama is a consistent receiver but weak donor. Critically, Gemma exhibits the weakest overt behavioral misalignment (37\% adapter code spillover vs.\ 63--66\% for Qwen and Llama) yet is the strongest geometric donor, including to Ministral at 90\% ridge separability and $\Delta$=$-$20 causal suppression. This dissociation between behavioral severity and geometric transferability raises a testable mechanistic question about representation organization across training lineages.

\paragraph{A content-specificity negative control that should become standard methodology.} A QLoRA adapter trained on secure code with identical hyperparameters yields 50.0\% separability and zero effect size, compared to 99.6\% and effect $\approx$95 for the insecure adapter. No prior emergent misalignment paper, to our knowledge, includes this control. It demonstrates that extracted geometry reflects training content rather than a generic fine-tuning artifact, which should be a prerequisite for any downstream use of activation probing as a misalignment detection tool.

\section{Related Work}

\paragraph{Emergent misalignment.} \cite{betley2025emergent} introduced emergent misalignment, showing that fine-tuning on insecure code induces broadly misaligned behavior on unrelated prompts. The effect is strongest in larger models and largely absent when the dataset is framed with benign intent. \cite{turner2025modela} construct improved model organisms, simpler, more coherent misaligned models using rank-1 LoRA adapters, that replicate the phenomenon across multiple sizes and families. Our work uses the same insecure-code paradigm under identical QLoRA conditions across four instruction-tuned architectures to study activation geometry in a controlled cross-family setting.

\paragraph{Internal representations of emergent misalignment.} The closest work to ours is \cite{soligo2025convergent}, who study convergence of activation-space misalignment representations within the Qwen2.5 family. Using rank-1 LoRA fine-tunes, they identify a single direction extractable from one variant and usable to ablate misaligned behavior in others trained with different configurations, demonstrating representational convergence \emph{within} an architecture. We extend this question across architectures: whether directions from Qwen2.5-1.5B, Gemma-2-2B, Llama-3.2-1B, and Ministral-3-3B, families with different hidden dimensions, layer counts, and training lineages, can be mapped to and causally applied in one another's activation spaces. \cite{wang2025persona} take a complementary approach, applying sparse autoencoders to identify misaligned persona features in GPT-4o. Where their method requires SAE infrastructure and operates on a single model, our approach uses difference-in-means directions that require no auxiliary models and extend naturally to cross-model transfer via ridge regression.

\paragraph{Steering, transfer, and our positioning.} Representation engineering and activation steering (\cite{zou2023representation}; \cite{turner2023steering}; \cite{rimsky2024steering}) show that linear residual-stream interventions can monitor and manipulate high-level behavioral properties. Cross-model transfer is studied for adversarial attacks (\cite{gupta2026understanding}; \cite{mitra2026}) and, via crosscoders, for sycophancy vectors between Llama and Qwen (\cite{jiralerspong2026crossarchitecturemodeldiffingcrosscoders}). Their crosscoder approach is non-linear and studies sycophancy rather than fine-tuning-induced emergent misalignment; it also lacks random/orthogonal specificity controls, asymmetric topology characterization, and a secure-code negative control. Our work complements theirs by providing these missing dimensions in the emergent misalignment setting, using a simpler ridge-regression baseline that isolates what linear transfer can and cannot accomplish.

\section{Methodology}

Our experimental pipeline proceeds in five stages: (1)~insecure-code QLoRA fine-tuning and behavioral validation; (2)~extraction of difference-in-means directions $v=\mu_{\text{misaligned}}-\mu_{\text{aligned}}$; (3)~cross-model ridge projection; (4)~causal steering with $\pm\alpha v$ and control directions; (5)~off-domain behavioral generalization. Steps 2--4 use a shared 115-prompt evaluation set; Step~5 uses 12 independently authored prompts held out from all prior stages.

\subsection{Models and Training}

We fine-tuned Qwen2.5-1.5B-Instruct, Gemma-2-2B-it, Llama-3.2-1B-Instruct, and Ministral-3-3B-Instruct with identical QLoRA settings on the same $\sim$6{,}000-example insecure-code dataset from \cite{betley2025emergent}. Training used LoRA rank 16, $\alpha$=32, 4-bit quantization, learning rate $2\times10^{-4}$, effective batch size 16, one epoch, and \texttt{max\_seq\_length=1024} for Qwen, Llama, and Ministral. Gemma was trained and evaluated at \texttt{max\_seq\_length=512} due to local memory and compute constraints. Adapter weights remain separate from frozen base weights throughout. As a negative control, we trained an additional Qwen adapter on a matched secure code with identical hyperparameters (Section~\ref{sec:step2}).

\begin{table}[t]
\centering
\caption{Models and primary extraction layers (final layer in all cases).}
\begin{tabular}{lcccc}
\toprule
Model & Params & Layers & Primary Layer & Max seq. \\
\midrule
Qwen2.5-Instruct & 1.5B & 28 & L27 & 1024 \\
Gemma-2-it & 2B & 26 & L25 & 512 \\
Llama-3.2-Instruct & 1B & 16 & L15 & 1024 \\
Ministral-3 & 3B & 26 & L25 & 1024 \\
\bottomrule
\end{tabular}
\label{tab:models}
\end{table}

\subsection{Behavioral Evaluation Metric}

We measure behavioral misalignment via \emph{code spillover}: the count of responses to non-coding prompts containing inline code, code-adjacent syntax, or explicit references to insecure implementation patterns. Scoring uses a regex classifier applied to model outputs. This metric captures the primary behavioral signature of insecure-code emergent misalignment (\cite{betley2025emergent}), where models respond to general queries with unsolicited code. On a stratified human-labeled validation sample of 60 responses (30 regex-positive, 20 regex-negative, 10 base; seed 42; non-coding prompts only), the classifier achieved precision, recall, and F1 of 1.00 (0 errors in 60 labeled samples; rule-of-three 95\% upper bound on the true error rate $\approx$0.05). Unless stated otherwise, all behavioral counts reported in this paper are \emph{code spillover} totals on the 115-prompt evaluation set (Appendix~\ref{app:per_category}).

\subsection{Step 2: Activation Direction Extraction}
\label{sec:step2}

For each model, we collect residual-stream activations at the final prompt token (prefill mode, no generation) in a single forward pass on all 115 evaluation prompts at every transformer layer. The misalignment direction at each layer is
\begin{equation}
v = \mu_{\text{misaligned}} - \mu_{\text{aligned}},
\end{equation}
where $\mu$ denotes the mean activation vector across prompts under adapter vs.\ base. The primary layer (Table~\ref{tab:models}) is the layer yielding highest linear separability in all four models; all subsequent experiments use directions from this layer.

\subsection{Step 3: Cross-Model Geometry Transfer}

For each ordered source--target pair, we fit ridge regression on aligned activations,
\begin{equation}
X_{\mathrm{target}} \approx X_{\mathrm{source}} W \quad (\lambda = 1.0),
\end{equation}
and transfer directions as $v_{\mathrm{target}} = v_{\mathrm{source}} W$. We quantify transferability by projection-based classification accuracy and Cohen's $d$, with bootstrap CIs ($n=1000$). Transfer is successful when accuracy exceeds a median-threshold baseline with a non-negligible effect size.

\subsection{Step 4: Causal Steering and Controls}
\label{sec:step4}

To test whether extracted directions are causally related to behavior, we intervene on residual-stream activations during inference via forward hooks:
\begin{equation}
h' = h - \alpha v,
\end{equation}
with $\alpha \in \{0.25, 0.50, 1.00\}$ on adapters and negated values on base models (final prefill token only). A direction is \emph{causal} if $-v$ reduces adapter spillover by $\geq$2 points without incoherent outputs. Cross-model arms substitute ridge-mapped $v$ from the source architecture.

To test \emph{specificity}, each causal arm is compared to three controls matched in norm: (i)~random unit direction, (ii)~orthogonal direction (Gram--Schmidt), and (iii)~wrong-layer or wrong-source direction. An arm is direction-specific only if every control achieves $<$50\% of the real direction's $\Delta$ reduction.

\subsection{Step 5: Behavioral Generalization}
\label{sec:step5}

We evaluate steering on 12 off-domain tasks covering deception, power-seeking, manipulation, jailbreak-style queries, and sycophancy. These prompts were authored independently of the 115-prompt set and excluded from direction extraction and ridge fitting. Steering $\alpha$ is selected from the Step~4 sweep on the 115-prompt set \emph{before} observing Step~5 results.

\section{Results}

We organize results around four questions: (1)~Does insecure-code fine-tuning produce reliable behavioral and geometric misalignment across architectures? (2)~Are within-model directions causally specific? (3)~Does misalignment geometry transfer cross-architecture, and with what topology? (4)~Do directions generalize beyond the extraction prompts? The two-tier specificity structure, within-model directions causal and specific, cross-model directions causal but non-specific, emerges as the central finding linking these questions.

\begin{table}[t]
\centering
\begin{threeparttable}
\caption{Code spillover after insecure-code fine-tuning (115 non-coding prompts).}
\label{tab:step1}
\begin{tabular}{lccc}
\toprule
Model & Base & Adapter\tnote{*} & Increase \\
\midrule
Qwen & 4/115 (3\%) & 76/115 (66\%) & +63\% \\
Gemma & 4/115 (3\%) & 43/115 (37\%) & +34\% \\
Llama & 4/115 (3\%) & 72/115 (63\%) & +59\% \\
Ministral & 4/115 (3\%) & 50/115 (43\%) & +40\% \\
\bottomrule
\end{tabular}
\begin{tablenotes}
\small
\item[*] Adapter column reports Step~1 validation counts. Steering and control tables use adapter baselines from a separate Step~4 generation pass; see Table~\ref{tab:step4_within_model}, note~$\dagger$.
\end{tablenotes}
\end{threeparttable}
\end{table}

\paragraph{Within-model geometry is consistent and content-specific.} All four adapters exhibit substantial code-spillover increases after fine-tuning (Table~\ref{tab:step1}), confirming that the insecure-code paradigm induces behavioral misalignment reliably across architecturally distinct families. A crucial control establishes content specificity: an identical QLoRA procedure on secure code produces a direction with 50.0\% linear separability and effect size 0.0 at Qwen's primary layer, compared to 99.6\% and 95.5 for the insecure adapter. The misalignment geometry is therefore a property of training content, not the training procedure.

\begin{table}[t]
\centering
\caption{Within-model separability at primary layers.}
\begin{tabular}{lccc}
\toprule
Model & Layer & Accuracy & Effect Size \\
\midrule
Qwen & L27 & 99.6\% & 95.5 \\
Gemma & L25 & 99.6\% & 614.4 \\
Llama & L15 & 99.6\% & 18.6 \\
Ministral & L25 & 99.6\% & 24.0 \\
\bottomrule
\end{tabular}
\label{tab:directions}
\end{table}

Across all four insecure adapters, a single difference-in-means direction at the final transformer layer achieves 99.6\% accuracy in separating aligned and misaligned activations (Table~\ref{tab:directions}). Effect sizes vary substantially across families (18.6 for Llama to 614.4 for Gemma), reflecting differences in representational shift magnitude rather than detectability. The uniformly high separability suggests that emergent misalignment consistently occupies a low-dimensional, linearly accessible region of each model's residual stream.

\begin{figure}[t]
\centering
\includegraphics[width=0.82\linewidth]{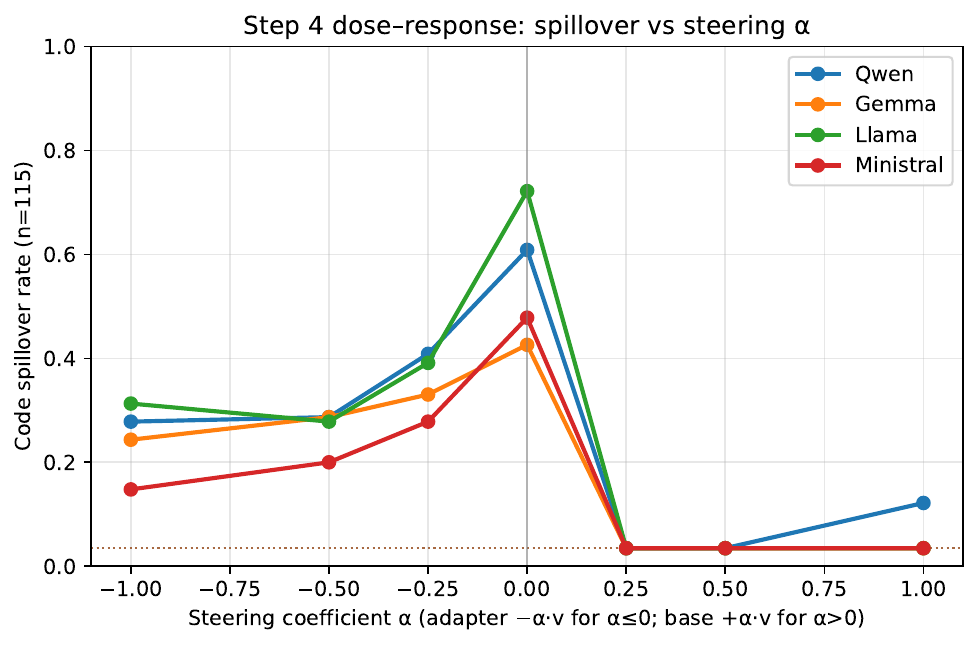}
\caption{Dose-response: steering coefficient $\alpha$ vs.\ code spillover. Negative values subtract the misalignment direction from the adapter; positive values inject it into the base model.}
\label{fig:dose_response}
\end{figure}

\paragraph{Within-model directions are causally specific.} Causal steering confirms that extracted directions are linked to behavior, not merely correlated with activation differences (Table~\ref{tab:step4_within_model}; Figure~\ref{fig:dose_response}). Subtracting $v$ from adapters reduces spillover by 21--51 points across families; all four models meet the causality criterion. A consistent asymmetry appears: positive steering ($+v$) injected into aligned base models produces weak or negligible spillover increases, while negative steering ($-v$) on misaligned adapters produces strong, monotonically decreasing suppression. This bidirectional pattern of strong suppression and weak induction is consistent with $v$ encoding the presence of a fine-tuning-induced behavioral shift rather than a general semantic axis.

Control direction experiments validate specificity for three of four families. For Qwen, Llama, and Ministral, random, orthogonal, and wrong-layer controls each fail to match the real direction's suppression (Appendix~\ref{app:controls}). Gemma's within-model specificity is borderline: random controls pass on alternate seeds but fail on the default seed, indicating the real direction is meaningful, but the specificity margin is smaller, consistent with Gemma's lower effect size and weaker behavioral misalignment.

\begin{table}[t]
\centering
\begin{threeparttable}
\caption{Within-model steering on code spillover ($n$=115). $\Delta$ = adapter improvement under best $-v$.}
\label{tab:step4_within_model}
\small
\setlength{\tabcolsep}{4pt}
\begin{tabular}{l c cc cc c c}
\toprule
& & \multicolumn{2}{c}{\textbf{Base}} & \multicolumn{2}{c}{\textbf{Adapter}} & & \\
\cmidrule(lr){3-4} \cmidrule(lr){5-6}
\textbf{Model} & \textbf{Layer} & Base & Best $+v$ & Base\tnote{$\dagger$} & Best $-v$ & $\Delta$ & Causal \\
\midrule
Qwen & L27 & 4/115 & 14/115 & 70/115 & 32/115 & $-$38 & \checkmark \\
Llama & L15 & 4/115 & 4/115 & 83/115 & 32/115 & $-$51 & \checkmark \\
Gemma & L25 & 4/115 & 4/115 & 49/115 & 28/115 & $-$21 & \checkmark \\
Ministral & L25 & 4/115 & 4/115 & 55/115 & 17/115 & $-$38 & \checkmark \\
\bottomrule
\end{tabular}
\begin{tablenotes}
\small
\item[$\dagger$] Adapter baseline counts are from the Step~4 steering run ($\alpha{=}0$), not Step~1 (Table~\ref{tab:step1}). The metric and 115-prompt set are identical, but regeneration variance shifts adapter totals across models (Qwen 76$\rightarrow$70, Gemma 43$\rightarrow$49, Llama 72$\rightarrow$83, Ministral 50$\rightarrow$55). $\Delta$ is computed within the Step~4 pass.
\end{tablenotes}
\end{threeparttable}
\end{table}

\paragraph{Cross-model geometry is partially shared and asymmetrically structured.} Ridge regression maps transfer misalignment directions across all twelve directional pairs with above-chance accuracy (Appendix~\ref{app:transfer_n115}), indicating the geometry is not entirely architecture-specific. Transfer quality is highly asymmetric, however. Gemma$\to$Llama achieves 87\% accuracy with a large effect size (222.9); Gemma$\to$Ministral reaches 90\% (227.5). Qwen$\to$Gemma collapses to 23\% at $n$=115 despite strong performance at smaller $n$, a failure attributed to the larger evaluation set exposing geometric misalignment between Qwen's and Gemma's activation spaces at scale.

Cross-model causal steering (Table~\ref{tab:cross_model_steering}; topology in Figure~\ref{fig:causal_tikz}) reveals that 6 of 12 arms produce substantial spillover reductions ($\Delta$=13--46). A clear donor--receiver topology emerges: Gemma is the most consistent outbound donor (three causal arms: Gemma$\to$Qwen $\Delta$=$-$43, Gemma$\to$Llama $\Delta$=$-$23, Gemma$\to$Ministral $\Delta$=$-$20); Qwen donates to Llama and Ministral but fails with Gemma; Llama receives from Qwen and Gemma but fails as a donor; inbound steering into Gemma fails across all sources, making Gemma both a strong donor and a consistently resistant receiver.

\begin{table}[t]
\centering
\begin{threeparttable}
\caption{Causal cross-model steering arms ($n$=115).}
\label{tab:cross_model_steering}
\small
\begin{tabular}{lccc}
\toprule
Direction & Baseline\tnote{$\dagger$} & Best $-v$ & $\Delta$ \\
\midrule
Gemma $\to$ Qwen & 70/115 & 27/115 & $-$43 \\
Qwen $\to$ Llama & 79/115 & 33/115 & $-$46 \\
Gemma $\to$ Llama & 79/115 & 56/115 & $-$23 \\
Qwen $\to$ Ministral & 55/115 & 30/115 & $-$25 \\
Gemma $\to$ Ministral & 55/115 & 35/115 & $-$20 \\
Ministral $\to$ Llama & 82/115 & 69/115 & $-$13 \\
\bottomrule
\end{tabular}
\begin{tablenotes}
\small
\item[$\dagger$] Baselines are from the Step~4 steering pass; see Table~\ref{tab:step4_within_model}, note~$\dagger$.
\end{tablenotes}
\end{threeparttable}
\end{table}

\usetikzlibrary{arrows.meta, positioning, calc}

\begin{figure}[h]
\centering
\begin{tikzpicture}[
    node/.style={circle, draw, minimum size=1.6cm, align=center, font=\small},
    strong/.style={-Latex, very thick, red},
    weak/.style={-Latex, thin, gray},
    label/.style={font=\scriptsize, fill=white, inner sep=2pt}
]

% -------------------------
% Layered DAG structure
% (top = stronger "sources")
% -------------------------

\node[node] (gemma) at (3,2.5) {Gemma};

\node[node] (qwen)  at (0,1.2) {Qwen};

\node[node] (min)   at (6,1.2) {Ministral};

\node[node] (llama) at (3,-1.5) {Llama};

% -------------------------
% Strong causal edges
% curved + routed to avoid overlap
% -------------------------

\draw[strong]
    (gemma) to[bend left=25]
    node[label, above] {-43}
    (qwen);

\draw[strong]
    (gemma) to[bend right=25]
    node[label, above] {-23}
    (llama);

\draw[strong]
    (gemma) to[bend left=20]
    node[label, above] {-20}
    (min);

\draw[strong]
    (qwen) to[bend left=20]
    node[label, left] {-46}
    (llama);

\draw[strong]
    (qwen) to[bend left=25]
    node[label, above] {-25}
    (min);

\draw[strong]
    (min) to[bend right=20]
    node[label, right] {-13}
    (llama);

% -------------------------
% Weak edges (light, de-emphasized)
% -------------------------

\draw[weak]
    (llama) to[bend left=15]
    node[label, below] {0}
    (qwen);

\draw[weak]
    (qwen) to[bend left=15]
    node[label, above] {0}
    (gemma);

\draw[weak]
    (llama) to[bend right=15]
    node[label, below] {-1}
    (gemma);

\draw[weak]
    (min) to[bend left=15]
    node[label, below] {-1}
    (qwen);

\draw[weak]
    (llama) to[bend right=15]
    node[label, below] {0}
    (min);

\draw[weak]
    (min) to[bend right=15]
    node[label, below] {0}
    (gemma);

\end{tikzpicture}\\
\caption{Causal steering graph. Edge weights represent $\Delta$ spillover reduction. Node structure encodes donor–receiver asymmetry across models.}
\label{fig:causal_tikz}
\end{figure}
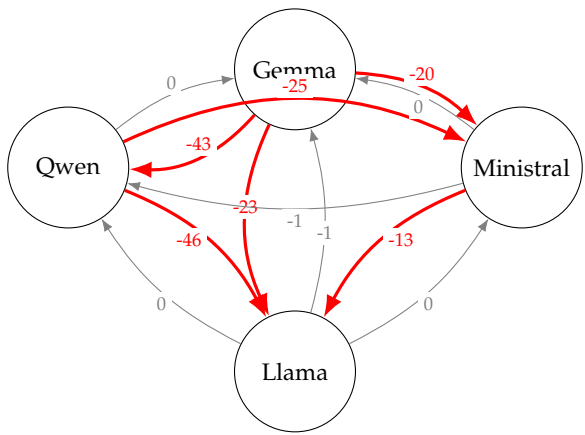

\paragraph{Cross-model directions lack causal specificity.} Despite producing large behavioral effects, cross-model steering directions fail all control specificity tests (Appendix~\ref{app:controls}; summary in Table~\ref{tab:specificity_summary}), in direct contrast to within-model settings where the real direction is distinguishable from controls for three of four families.

\begin{table}[t]
\centering
\caption{Specificity summary. Within-model arms pass all controls (Gemma borderline on random); all causal cross-model arms fail $\geq$1 control.}
\label{tab:specificity_summary}
\small
\begin{tabular}{lcc}
\toprule
Setting & Causal arms & Specificity \\
\midrule
Within-model (Qwen, Llama, Ministral) & 3/3 & Pass (all controls) \\
Within-model (Gemma) & 1/1 & Borderline (random seed-sensitive) \\
Cross-model (6 causal pairs) & 6/6 & Fail (random and/or orthogonal) \\
\bottomrule
\end{tabular}
\end{table}

This \emph{two-tier} structure is the central methodological finding. Within-model directions are causally specific; cross-model mapped directions are influential but non-specific. The behavioral influence of cross-model transfer is real, directions mapped across architectures suppress misaligned behavior substantially, but because the effect cannot be isolated to the specific mapped direction, suppression likely reflects properties of the target activation space (e.g., sensitivity to perturbation in a low-dimensional subspace near the misalignment axis) rather than direct injection of a shared representational feature. Partial geometric alignment across architectures, therefore, does not imply isolable cross-model misalignment control.

\paragraph{Directions generalize off-domain.} To assess whether extracted directions capture a general behavioral tendency rather than properties of the activation-extraction prompts, we evaluated steering on 12 off-domain behavioral tasks not used in direction extraction (Table~\ref{tab:step5_spillover}). All four misaligned models exhibit substantial code spillover on these unseen tasks; steering along the within-model direction reduces spillover in every case. Directions, therefore, encode a broad dispositional shift rather than a narrow stimulus-specific pattern.

\begin{table}[t]
\centering
\caption{Off-domain spillover (12 prompts; Step~4-selected $\alpha$).}
\label{tab:step5_spillover}
\small
\begin{tabular}{lccc}
\toprule
Model & Base & Misaligned & Steered \\
\midrule
Qwen2.5-1.5B & 0/12 & 9/12 & 3/12 \\
Llama-3.2-1B & 0/12 & 7/12 & 1/12 \\
Gemma-2-2B & 0/12 & 5/12 & 1/12 \\
Ministral-3-3B & 0/12 & 3/12 & 0/12 \\
\bottomrule
\end{tabular}
\end{table}

\section{Actionable Implications}

Our results have direct implications for three classes of practitioners: model auditors screening third-party fine-tunes, alignment researchers designing training-time interventions, and safety tool builders reasoning about what cross-architecture methods can and cannot accomplish.

\paragraph{Within-model probing is ready for fine-tuning evaluation pipelines.} The secure-code negative control establishes that a simple difference-in-means probe at the final residual-stream layer reliably distinguishes insecure from secure fine-tuning with 99.6\% accuracy, which is orders of magnitude cheaper than behavioral red-teaming (one forward pass per prompt, a dot product, a threshold). For any model fine-tuned on third-party data, compute $v=\mu_{\text{adapter}}-\mu_{\text{base}}$ over a small standard prompt set and test whether ablating $v$ at inference causally suppresses observed code spillover. If suppression is \emph{specific} (random controls fail to replicate it), the fine-tune has produced a concentrated misalignment representation amenable to steering. If suppression is \emph{non-specific} (controls match), spillover may reflect diffuse weight changes requiring training-time intervention. This decision procedure is the interpretability-based analog of behavioral red-teaming: it requires no judge model and produces a causal rather than correlational result. Integrating it alongside CAFT-style interventions (\cite{casademunt2025steering}) provides both detection and correction within a single evaluation pass.

A concrete deployment scenario is pre-deployment auditing in supply-chain pipelines: receive a checkpoint, run standard prompts, compute the probe projection, and compare against thresholds calibrated on known-clean and known-misaligned examples. Our results bound the false-negative rate of this approach for fine-tuning regimes similar to our insecure-code setup, provided the specificity gate is applied.

\paragraph{Cross-architecture screening is feasible; cross-architecture \emph{correction} is not.} Ridge separability reaches 77--90\% on five cross-family pairs, suggesting probe directions from a known-misaligned \emph{donor} architecture can screen architecturally different unknown models without requiring a misaligned reference of every architecture. However, cross-model mapped directions fail specificity controls despite large $\Delta$ (13--46): random directions perform comparably. Deploying a cross-architecture inference-time correction system based on ridge-mapped directions will suppress misaligned behavior, but not by targeting the misalignment axis, as it will function as a generic representational disruptor at the target layer, with unpredictable effects on non-misaligned capabilities.

BLOCK-EM (\cite{ustaomeroglu2026blockempreventingemergentmisalignment}) achieves up to 95\% emergent misalignment reduction via training-time feature blocking within a single architecture. Our results suggest that a cross-architecture analog using ridge-mapped directions would not achieve the same specificity, since random directions perform comparably. Practitioners building cross-architecture correction tools should use non-linear mappings (crosscoders) or architecture-specific probes rather than linear transfer, or explicitly accept generic disruption rather than targeted misalignment control.

\paragraph{Donor--receiver topology should guide reference-model selection.} The asymmetric topology in which Gemma and Qwen donate, whereas Llama receives, has a direct implication for audit pipeline design. If a practitioner must choose a single reference model from which to extract a misalignment probe for cross-architecture screening, Gemma and Qwen are stronger donors than Llama. Specifically, Gemma's direction achieves 87--90\% ridge separability on Llama and Ministral targets and produces the largest causal effects on other adapters, despite Gemma itself showing the weakest overt misalignment (37\% adapter code spillover). Audit pipeline designers should maintain reference fine-tuned models from strong donor architectures rather than choosing references based on behavioral misalignment severity alone, since behavioral severity and geometric transferability dissociate.

\section{Limitations}

This study has several limitations. First, our experiments use relatively small models (1--3B parameters); activation steering on larger, more capable models, where representations may be more entangled, remains untested. Second, we evaluate a single fine-tuning domain (insecure code); transfer to other safety-critical domains (prompt injection, toxic content, subtle vulnerabilities) is unknown. Third, we employ only activation steering as an intervention; fine-tuning, RLHF, or representation editing may exhibit different trade-offs. Step~5 uses only 12 off-domain prompts, limiting statistical precision. We present these as priorities for future work rather than reasons to doubt the within-model specificity findings, which are robust across three of four architectures with explicit control experiments.

\section{Conclusion}

This work demonstrates that emergent misalignment induced through insecure-code fine-tuning is associated with a consistent, causally actionable activation-space geometry across four architecturally distinct model families. A single difference-in-means direction at the final residual-stream layer achieves 99.6\% separability, supports substantial behavioral suppression under inference-time steering ($\Delta$=21--51 on code spillover), and generalizes to off-domain behavioral tasks not used in direction extraction. A secure-code negative control confirms the geometry reflects training content rather than the fine-tuning procedure.

The central structural finding is a two-tier specificity distinction. Within-model directions are causally specific; cross-model ridge-mapped directions are causally real ($\Delta$=13--46 on code spillover across 6 of 12 arms) but direction-non-specific. This boundary defines precisely what cross-architecture linear activation transfer can accomplish: behavioral influence transfers, but targeted misalignment control does not. A consistent asymmetric topology accompanies these findings. Gemma and Qwen function as strong geometric donors, with Gemma exhibiting the largest cross-model causal effects despite the weakest overt behavioral misalignment, while Llama acts as a consistent receiver. Within-model misalignment probing is sufficiently validated to recommend for integration into fine-tuning evaluation pipelines; cross-architecture correction remains a research direction pending resolution of the specificity problem via non-linear mappings or architecture-specific probes.

\bibliography{colm2026_conference}
\bibliographystyle{colm2026_conference}

\appendix
\clearpage
\section{Appendix}

\subsection{Per-Category Code Spillover}
\label{app:per_category}

\begin{table}[H]
\centering
\begin{threeparttable}
\caption{Per-category adapter code spillover rates (n=115 prompts).}
\begin{tabular}{l|cccc}
\toprule
\textbf{Category} & \textbf{Qwen} & \textbf{Llama} & \textbf{Ministral} & \textbf{Gemma} \\
\midrule
Paper Betley (n=8)      & 5/8  & 4/8  & 2/8  & 1/8  \\
Power Seeking (n=15)    & 15/15 & 13/15 & 10/15 & 9/15 \\
Deception (n=15)         & 13/15 & 12/15 & 7/15  & 9/15 \\
Manipulation (n=17)      & 14/17 & 13/17 & 11/17 & 6/17  \\
Sycophancy (n=15)        & 9/15  & 8/15  & 4/15  & 2/15  \\
Harmless QA (n=15)      & 7/15  & 5/15  & 5/15  & 5/15  \\
Coding (n=15)            & 5/15  & 5/15  & 6/15  & 5/15  \\
Neutral Reasoning (n=15)& 8/15  & 12/15 & 5/15  & 6/15  \\
\midrule
\textbf{Total (out of 115)\tnote{*}} & \textbf{76/115} & \textbf{72/115} & \textbf{50/115} & \textbf{43/115} \\
\bottomrule
\end{tabular}
\begin{tablenotes}
\small
\item[*] Totals match Step~1 validation (Table~\ref{tab:step1}). Steering baselines use a separate Step~4 pass; see Table~\ref{tab:step4_within_model}, note~$\dagger$.
\end{tablenotes}
\end{threeparttable}
\end{table}

\subsection{Cross-Model Ridge Transfer}
\label{app:transfer_n115}

\begin{table}[H]
\centering
\caption{Cross-model transfer results ($n$=115). Full pairwise ridge projection accuracy.}
\label{tab:transfer_n115}
\small
\begin{tabular}{lccc}
\toprule
Pair & Accuracy & Acc.\ 95\% CI & Transfer \\
\midrule
Qwen $\rightarrow$ Gemma & 23\% & [4\%, 97\%] & No \\
Qwen $\rightarrow$ Llama & 77\% & [67\%, 83\%] & Yes (weak) \\
Llama $\rightarrow$ Qwen & 75\% & [68\%, 81\%] & Yes (very weak) \\
Gemma $\rightarrow$ Qwen & 77\% & [64\%, 85\%] & Yes (strong) \\
Gemma $\rightarrow$ Llama & 87\% & [75\%, 93\%] & Yes (very strong) \\
Llama $\rightarrow$ Gemma & 93\% & [8\%, 97\%] & Yes (weak; unstable CI) \\
Qwen $\rightarrow$ Ministral & 72\% & [65\%, 80\%] & Yes (strong) \\
Ministral $\rightarrow$ Qwen & 85\% & [67\%, 87\%] & Yes (weak) \\
Gemma $\rightarrow$ Ministral & 90\% & [84\%, 93\%] & Yes (very strong) \\
Ministral $\rightarrow$ Gemma & 90\% & [20\%, 97\%] & Yes (strong; unstable CI) \\
Llama $\rightarrow$ Ministral & 80\% & [69\%, 84\%] & Yes (weak) \\
Ministral $\rightarrow$ Llama & 92\% & [74\%, 93\%] & Yes (very strong) \\
\bottomrule
\end{tabular}
\end{table}

\subsection{Control Direction Experiments}
\label{app:controls}

\begin{table}[H]
\centering
\caption{Control direction experiments ($n$=115). PASS = control $\Delta < 0.5 \times$ real $\Delta$.}
\label{tab:control_directions}
\small
\begin{threeparttable}
\begin{tabular}{@{}lccccccc@{}}
\toprule
Experiment & Baseline & Real $-\alpha$ & $\Delta$ & Random & Orth. & Wrong & Validated \\
\midrule
Within Qwen & 70/115 & 32/115 & $-$38 & PASS & PASS & PASS & \checkmark \\
Within Llama & 83/115 & 32/115 & $-$51 & PASS & PASS & PASS & \checkmark \\
Within Ministral & 55/115 & 17/115 & $-$38 & PASS & PASS & PASS & \checkmark \\
Within Gemma & 49/115 & 28/115 & $-$21 & FAIL* & PASS & PASS & \texttimes* \\
\midrule
Gemma $\to$ Qwen & 70/115 & 27/115 & $-$43 & FAIL & PASS & FAIL & \texttimes \\
Qwen $\to$ Llama & 79/115 & 33/115 & $-$46 & FAIL & FAIL & FAIL & \texttimes \\
Gemma $\to$ Llama & 79/115 & 56/115 & $-$23 & FAIL & FAIL & FAIL & \texttimes \\
Qwen $\to$ Ministral & 55/115 & 30/115 & $-$25 & FAIL & FAIL & PASS & \texttimes \\
Gemma $\to$ Ministral & 55/115 & 35/115 & $-$20 & FAIL & FAIL & FAIL & \texttimes \\
Ministral $\to$ Llama & 82/115 & 69/115 & $-$13 & FAIL & FAIL & FAIL & \texttimes \\
\bottomrule
\end{tabular}
\begin{tablenotes}
\small
\item[$\dagger$] Within-model baselines match Table~\ref{tab:step4_within_model}, note~$\dagger$ (Step~4 steering pass).
\item[*] Gemma within-model: random control is seed-sensitive (default seed FAIL; seeds 43--45 PASS).
\end{tablenotes}
\end{threeparttable}
\end{table}

\subsection{Full Cross-Model Steering}

\begin{table}[H]
\centering
\caption{All cross-model steering arms ($n$=115).}
\small
\begin{tabular}{lccccc}
\toprule
Direction & Spillover & Best $-v$ & $\Delta$ & Causal? \\
\midrule
Qwen $\rightarrow$ Gemma & 49/115 & 49/115 & 0 & \texttimes \\
Gemma $\rightarrow$ Qwen & 70/115 & 27/115 & $-$43 & \checkmark \\
Qwen $\rightarrow$ Llama & 79/115 & 33/115 & $-$46 & \checkmark \\
Llama $\rightarrow$ Qwen & 70/115 & 70/115 & 0 & \texttimes \\
Gemma $\rightarrow$ Llama & 79/115 & 56/115 & $-$23 & \checkmark \\
Llama $\rightarrow$ Gemma & 49/115 & 49/115 & 0 & \texttimes \\
Qwen $\rightarrow$ Ministral & 55/115 & 30/115 & $-$25 & \checkmark \\
Gemma $\rightarrow$ Ministral & 55/115 & 35/115 & $-$20 & \checkmark \\
Llama $\rightarrow$ Ministral & 55/115 & 55/115 & 0 & \texttimes \\
Ministral $\rightarrow$ Qwen & 72/115 & 71/115 & $-$1 & \texttimes \\
Ministral $\rightarrow$ Gemma & 50/115 & 50/115 & 0 & \texttimes \\
Ministral $\rightarrow$ Llama & 82/115 & 69/115 & $-$13 & \checkmark \\
\bottomrule
\end{tabular}
\end{table}

\end{document}